\documentclass[english,runningheads,a4paper]{llncs}
\usepackage[utf8]{inputenc}
\usepackage[T1]{fontenc}
\usepackage{graphicx}
\usepackage{grffile}
\usepackage{longtable}
\usepackage{wrapfig}
\usepackage{rotating}
\usepackage[normalem]{ulem}
\usepackage{amsmath}
\usepackage{textcomp}
\usepackage{amssymb}
\usepackage{capt-of}
\usepackage{hyperref}
\usepackage{booktabs}\usepackage{multirow}\usepackage{adjustbox}
\date{\today}
\title{SlideImages: A Dataset for \\
Educational Image Classification}
\author{David Morris\inst{1} \and Eric Müller-Budack\inst{1} \and Ralph Ewerth\inst{1}\inst{2}}
\institute{TIB -- Leibniz Information Centre for Science and Technology, Hannover, Germany,\\
\email{\{David.Morris, Eric.Mueller, Ralph.Ewerth\}@tib.eu}
\and L3S Research Center, Leibniz Universit\"at Hannover, Hannover, Germany}
\hypersetup{
 pdfauthor={},
 pdftitle={SlideImages: A Testing Dataset for Educational Image Classification},
 pdfkeywords={},
 pdfsubject={},
 pdfcreator={Emacs 26.3 (Org mode 9.1.14)}, 
 pdflang={English}}
\begin{document}

\maketitle
\begin{abstract}
In the past few years, convolutional neural networks (CNNs) have achieved impressive results in computer vision tasks, which however mainly focus on photos with natural scene content. Besides, non-sensor derived images such as illustrations, data visualizations, figures, etc. are typically used to convey complex information or to explore large datasets. However, this kind of images has received little attention in computer vision. 
CNNs and similar techniques use large volumes of training data. Currently, many document analysis systems are trained in part on scene images due to the lack of large datasets of educational image data.
In this paper, we address this issue and present SlideImages, a dataset for the task of classifying educational illustrations. SlideImages contains training data collected from various sources, e.g., Wikimedia Commons and the AI2D dataset, and test data 
collected from educational slides. We have reserved all the actual educational images as a test dataset in order to ensure that the approaches using this dataset generalize well to new educational images, and potentially other domains. Furthermore, we present a baseline system using a standard deep neural architecture and discuss dealing with the challenge of limited training data.
\end{abstract}
\section{Introduction}
\label{sec:org8c5fa32}

Convolutional neural networks (CNNs) are making great strides in computer vision, driven by large datasets of annotated photos, such as ImageNet \cite{imagenet_cvpr09}. Many images relevant for information retrieval, such as charts, tables, and diagrams, are created with software rather than through photography or scanning. 

There are several applications in information retrieval for a robust classifier of educational illustrations. Search tools might directly expose filters by predicted label, natural language systems could choose images by type based on what information a user is seeking. Further analysis systems could be used to extract more information from an image to be indexed based on its class. In this case, we have classes such as pie charts and x-y graphs that indicate what type of information is in the image (e.g., proportions, or the relationship of two numbers) and how it is symbolized (e.g., angular size, position along axes).

Most educational images are created with software and
are qualitatively different from photos and scans.
Neural networks designed and trained to make sense of the noise and spatial relationships in photos are sometimes suboptimal for born-digital images and educational images in general.

Despite their practical relevance, educational images and illustrations have been comparatively underserved in training datasets and challenges. Competitions such as the
Contest on Robust Reading for Multi-Type Web Images \cite{mtwiContest} and ICDAR
DeTEXT \cite{detextContest} have shown that these tasks are difficult and
unsolved. Research on text extraction such as Morris et al. \cite{self} and Nayef
\& Logier \cite{frenchLocalization} has shown that even noiseless born-digital
images are sometimes better analyzed with neural nets than with handcrafted
features and heuristics. Born-digital and educational images need further benchmarks related to challenging information
retrieval tasks in order to test the generalization of methods for those tasks.

In this paper, we propose SlideImages, a dataset which targets images from educational presentations. Most of these educational illustrations are created with diverse software, so the same symbols are drawn in different ways in different parts of the image. As a result, we expect that effective synthetic datasets will be hard to create, and methods effective on SlideImages will generalize well to other tasks with similar symbols. SlideImages contains eight classes of image types (e.g. bar charts and x-y plots) and a class for photos. The labels we have created were made with information extraction for image summarization in mind.

In the rest of this paper, we discuss related work in §2, details about
our dataset and baseline method in §3, results of our baseline method in §4, and conclude with a discussion of potential future developments in §5.
\section{Related Work}
\label{sec:orgd0b8d5a}

Prior information retrieval publications have used, or  could make use of, document figure classification. Charbonnier et al. \cite{luciaECIR} built
a search engine which allows user to filter images based on type, Aletras \&
Mittal \cite{DBLP:conf/ecir/AletrasM17} automatically
label topics in photos, Kembhavi et al. \cite{diagram} build a system for extracting the relationships between entities in a diagram, which assumes that the input figure is a diagram, Hiippala \& Orekhova extended their dataset by annotating it semantically in terms of Relational Structure Theory, which implies that the same visual features communicate the same semantic relationships, and de Herrera
et al. \cite{DBLP:conf/ecir/HerreraMJSFM15} seek to classify image types to filter
their search for medical professionals.

We intend to use document figure classification as a first step in automatic educational image summarization applications. A similar idea is followed by Morash et
al. \cite{DBLP:journals/taccess/MorashSMHL15}, who built one template for each type of image, then manually classified images and filled out
the templates, and suggested automating the steps of that
process. Moraes et al. \cite{DBLP:conf/assets/MoraesSMC14} mentioned the same idea for their SIGHT (Summarizing Information GrapHics Textually) system.

A number of publications on document image classification
such as Afzal et al. \cite{errorByHalf} and Harley et al. \cite{rvl-cdip} use the RVL-CDIP (Ryerson Vision Lab Complex Document Information Processing) dataset, which
covers scanned documents.  While document scans and born-digital educational
illustrations have materially different appearance, these papers show that
the utility of deep neural networks is not limited to scene image tasks.

A classification dataset of scientific illustrations was created for the NOA project \cite{luciaTPDL}.
However, their dataset is not publicly available, and does not draw as
many distinctions between types of educational illustrations. 

Jobin et al.'s DocFigure \cite{docfigure} consists of 28 different categories of illustrations
extracted from scientific publications totaling 33,000 images. 
\section{Dataset and Baseline System}
\label{sec:orgaa965cd}
Techniques that work well on DocFigure \cite{docfigure} do not generalize to the educational illustrations in our use case scenarios (as we also show in section 4.2).
This suggests a dataset of specifically
educational illustrations is needed.

 CNNs and related techniques are heavily data driven. An approach must consist of both an architecture and optimization technique, but also the data used for that optimization.  In our case, we consider the dataset  our main contribution.
\subsection{SlideImages Dataset}
\begin{table}[!t]
\caption{\label{tab:org783f50b}
Dataset sizes, including reduced datasets for head-to-head comparison}
\centering
\begin{tabular}{lrrrr}
 & Classes & Train & Val & Test\\
\hline
SlideImages & 9 & 2646 & 292 & 691\\
DocFigure & 28 & 19795 & 0 & 13172\\
Head-to-Head SlideImages & 8 & 2331 & 257 & 575\\
Head-to-Head DocFigure & 8 & 11678 & 3886 & 3891\\
\end{tabular}
\end{table}

\begin{figure}[!t]
\centering
\includegraphics[width=.6\linewidth,trim=0 0 0 0, clip]{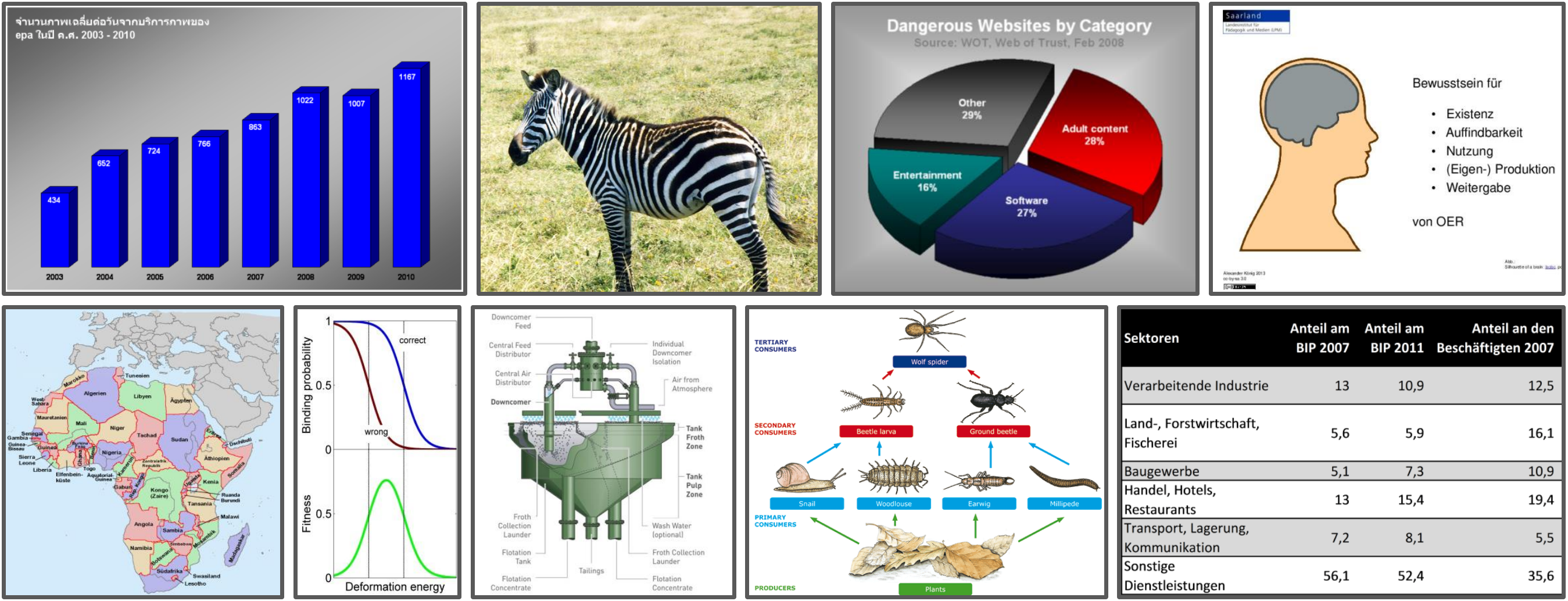}
\caption{\label{fig:dataset_ex}
Examples of our classes from training data. Clockwise from top left: bar charts, photos, pie charts, slide images, tables, structured diagrams, technical drawings, x-y plots, and maps.}
\end{figure}

When building our taxonomy, we have chosen classes such that one class would have the same types of salient features, and appropriate summaries would also be similar in structure. Our classes are also all common in educational materials. Beyond the requirements of our taxonomy, our datasets needed to be representative of common educational illustrations in order to fit real-world applications. We also aimed to create a dataset which we could legally re-share in order to promote research on the classification of educational images.

Educational illustrations are created by a variety of
communities with varying expertise, techniques, and tools, so choosing a
dataset from one source may eliminate certain variables in educational
illustration.  
To identify these variables, we kept our training and test data sources separate.

We have assembled a training dataset from various sources of open access
illustrations. Bar charts, x-y plots, maps, photos, pie charts, slide images,
table images, and technical drawings have been manually selected from Wikimedia
Commons images using the Wikimedia Commons search for related terms, and
choosing a varied group of suitable images. Graph diagrams, which we also call node-edge diagrams or ``structured diagrams,'' have been selected manually from the AllenAI
Diagram Understanding (AI2D) dataset described by Kembhavi et al. \cite{ai2d},
since not all of the diagrams in AI2D contain graph edges. AI2D was an ideal source of images for SlideImages, since Kembhavi et al. chose images which represented topics from grade school science \cite{ai2d}.

The test dataset of SlideImages is derived from a snapshot of SlideWiki open
educational resource platform (https://slidewiki.org/) datastore obtained in 2018.  From that snapshot, we manually selected and labeled 691 images.
We will make our training and test datasets available online and link them in
the camera-ready version of this paper.

\subsection{Baseline Approach}
\label{sec:methodology}
The SlideImages training dataset is small compared to datasets like ImageNet \cite{imagenet_cvpr09}, with over 14 million images, RVL-CDIP \cite{rvl-cdip} with 400,000 images, or even DocFigure \cite{docfigure} with 33,000 images.  Much of our methodology is shaped by needing to confront the challenges of a small dataset. In particular, we aim to avoid overfitting: the tendency of a classifier to identify individual images and patterns specific to the training set rather than the desired semantic concepts.

For our pre-training dataset, a large, diverse dataset is required that contains a large proportion of educational and scholarly images. 
We pre-trained on a dataset of almost 60,000 images labeled by Sohmen et al.~\cite{luciaECIR} (NOA dataset), provided by the authors on request. The images are
categorized as composite images, diagrams, medical imaging, photos, or visualizations/models.

We used image augmentation to help combat overfitting.  Distorting an image in ways which do not remove the characteristic qualities helps to disrupt spurious patterns that the network might otherwise pick up.  We used image stretching, brightness scaling, zooming, and color channel shifting, with details shown in our source code.  We also added dropout  with a rate of 0.1 on the extracted features before running our fully connected and output layers.  We used similar image augmentation for pre-training and training.

We use MobileNetV2 \cite{mobilenetv2} as our network architecture. We chose MobileNetV2 since it provides a balance between a small number of
parameters and proven performance on ImageNet. Intuitively, a smaller parameter space implies a model with more bias and lower variance, which is better for smaller datasets.  We initialized our
weights from an ImageNet model and pre-trained for a further 40 epochs
with early stopping on the NOA dataset using the Adam (adaptive moment estimation) \cite{adam} optimizer.  This additional pre-training was intended to cause the lower levels of the network to extract more features specific to born-digital images. We then trained for 40 epochs with Adam and a learning rate schedule. 
Our schedule drops the learning rate by a factor of 10 at the 15th and 30th epoch.

\section{Preliminary Results}
\label{sec:org604bce1}

\begin{figure}[!t]
\centering
\includegraphics[width=.5\linewidth,trim=55 168 355 60, clip]{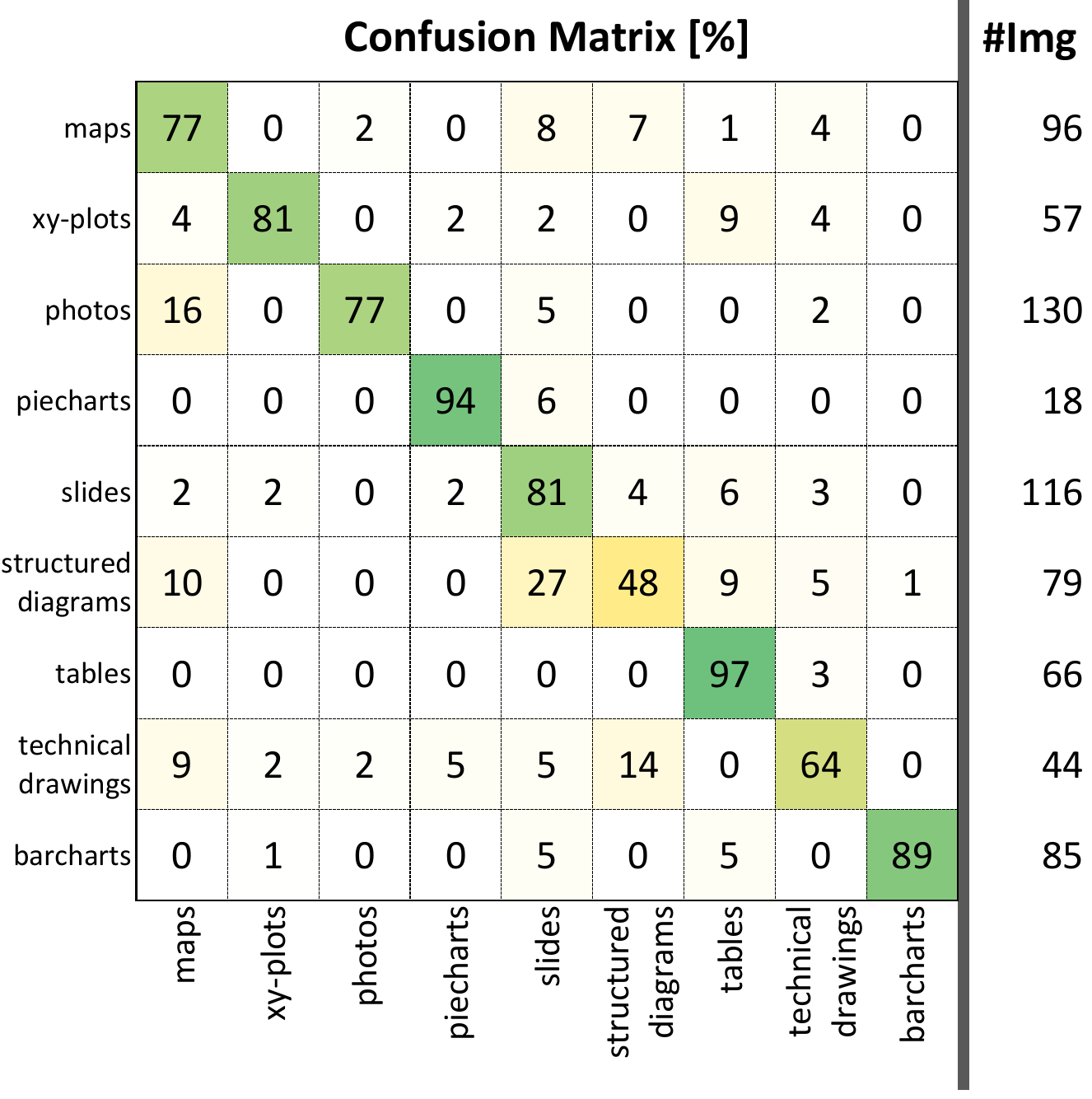}
\caption{\label{fig:org06abea2}
Confusion matrix of our baseline system on SlideImages. Entries show percent of true members of the class on the left margin labeled as on the bottom margin. Weighted accuracy average is 80\% over all 691 images.}
\end{figure}


We have performed two experiments, in order to show that this dataset represents
a meaningful improvement over existing work, and to establish a baseline. Because our classes are unbalanced, we have reported summary statistics as accuracy averages of each class weighted by number of instances per class.

\subsection{Baseline}
\label{sec:orgc7567cb}
We have used the classifier described in section \ref{sec:methodology} to generate a
baseline for our dataset. 
The confusion matrix in Fig. \ref{fig:org06abea2} shows that while
misclassifications do tend towards a few types of errors, none of the classes
have collapsed; while certain classes are likely to be misclassified as another
specific class (such as structured diagrams as slides), those relationships
do not happen in reverse, and a correct classification is more likely. Fig. \ref{fig:org06abea2} shows that our baseline leaves room for improvement, and our test set helps to identify challenges in this task. Viewing individual classification errors of our baseline highlighted a few
problems with our existing training data. Our training data do not include
sufficient structured diagrams with wide, illustrated arrows, or edges which
travel only at 90° increments, such as organigrams charts or many types of UML (Unified Modeling Language) diagrams.
Our photos do not include examples with the background removed, but these are
common in educational images. These problems should be remedied in future
training datasets in this and similar problems.

\begin{table}[htbp]
\caption{\label{tab:org9d01920}
Head-to-head comparison of precision weighted averages.}
\centering
\setlength{\tabcolsep}{6pt}
\begin{tabular}{lccc}
 & SlideImages Train & DocFigure Train & DocFigure Baseline\\
SlideImages Test & 80\% & 78\% & 75\%\\
DocFigure Test & 92\% & 99\% & 99\%\\
\end{tabular}
\end{table}

\subsection{Head-to-head Comparison}
\label{sec:org9b8ff98}
The related DocFigure dataset covers similar images and has much more data than
SlideImages.  To
justify SlideImages, we have created a head-to-head comparison of classifiers
trained in the same way (as described in section \ref{sec:methodology}) on the
SlideImages and DocFigure datasets. All the SlideImages
classes except \emph{slides} have an equivalent in DocFigure. We have shown the reduction in the data used, and the relative sizes of
the datasets, in Table \ref{tab:org783f50b}; the Head-to-Head datasets contain only the matching classes, and in the case of the DocFigure dataset,
the original test set has been split into validation and test sets.

After obtaining the two trained networks, we have tested each network on both
the matching testing set, and the other testing set. Although we were unable to reproduce the VGG-V baseline used by Jobin et al., we used a linear SVM with VGG-16 features and achieved comparable results on the full DocFigure dataset (90\% macro average compared to their 88.96\% with a fully neural feature extractor). We show our results in
Table \ref{tab:org9d01920}. The results show that SlideImages is a more challenging and
potentially more general task; the net trained on SlideImages did even better on
the DocFigure test set than on the SlideImages test set. Despite having a different source and
approximately a fifth of the size of the DocFigure dataset, our training set was
better on our test set.

\section{Conclusions and Future Work}
\label{sec:org9b50a5d}
In this paper, we have presented the task of classifying educational illustrations and images in slides and introduced a novel dataset SlideImages for this task. The classification remains an open problem despite our baseline and represents a useful task for information retrieval. We
have provided a test set derived from actual educational illustrations, and a
training set compiled from open access images. Finally, we have established a baseline system for the classification task. 

Other potential avenues for future research include experimenting with the DocFigure
dataset in the pre-training and training phases, and experimenting with text
extraction for multimodal classification.
\section{Acknowledgement}
This work is financially supported by the German Federal Ministry of Education and Research (BMBF) and European Social Fund (ESF) (InclusiveOCW project, no. 01PE17004).
\bibliographystyle{ieeetr}
\bibliography{classification}
\end{document}